# Navigating Multilingual News Collections Using Automatically Extracted Information

Ralf Steinberger, Bruno Pouliquen, Camelia Ignat
*European Commission – Joint Research Centre*
*Via E. Fermi 1, T.P. 267, 21020 Ispra (VA), Italy,*
http://www.jrc.it/langtech, http://press.jrc.it/NewsExplorer – Firstname.Lastname@jrc.it

**Abstract**. *We are presenting a text analysis tool set that allows analysts in various fields to sieve through large collections of multilingual news items quickly and to find information that is of relevance to them. For a given document collection, the tool set automatically clusters the texts into groups of similar articles, extracts names of places, people and organisations, lists the user-defined specialist terms found, links clusters and entities, and generates hyperlinks. Through its daily news analysis operating on thousands of articles per day, the tool also learns relationships between people and other entities. The fully functional prototype system allows users to explore and navigate multilingual document collections across languages and time.*

## 1. Introduction

Most large organisations employ full-time personnel to search the media for relevant information: Politicians and political parties are interested in the media's opinions expressed about them or about certain subjects; Organisations with an interest in medical or financial issues monitor reports about the outbreak of diseases, about stock exchange-relevant transactions, etc. The various users have in common that they need to sieve through vast amounts of news articles or other reports every day, in order to find the most relevant documents, to digest and store them. When analysts search this manually compiled text repository, they are again confronted with the time-consuming and monotonous task of having to look through large document collections.

The text analysis tool set presented here aims at improving the efficiency of this process significantly, by ordering any collection into related documents, by extracting information from each cluster of related documents, by visualising the results in an intuitive manner, and by allowing the information seekers to explore the document collection via automatically generated hyperlinks.

In this paper, we start from the assumption that the organisation already is in possession of the document collection, because they purchase news wires from press agencies or because they receive lists of potentially relevant articles from a news monitoring system like the *Europe Media Monitor* EMM ([1]). News is often distributed in the standardised XML format RSS ([6]). We therefore take it for granted that the documents are clean and well structured, or that separate format conversion tools convert other formats like HTML or PDF into RSS.

The following sections describe the extraction of named entities and terms from news texts (Sect. 2), as well as the clustering of similar documents and the monolingual and cross-lingual linking of related clusters (Sect. 3). Sect. 4 explains how the system learns relationships between entities over time. Sect. 5 presents the usage scenario by listing some ways of exploring an existing text collection with this tool. We end with a summary and by pointing to future work (Sect. 6).

## 2. Extraction of information from texts

The contents of factual news can roughly be summarised as *Who does What to Whom, Where and When*. Our aim is to cluster news articles that talk about the same event or story into one group (Sect. 3), and to then provide the users with answers to as many of these questions as possible: persons and organisations involved (Who, to Whom), keywords and specialist terminology found in the texts (What), the geographical coverage of a news story (Where), and the period of time concerned (When). Named Entity

Recognition is a rather well-known field of text analysis. For an overview of the state-of-the-art, see, for instance, [2].

## 2.1. Named entities

We store all the information that has been extracted from articles in a relational database. For the recognition of named entities, we do not use part-of-speech taggers or parsers because we aim at analysing all 20 official European Union (EU) languages and it is difficult and expensive to procure such linguistic tools for all these languages. Instead, we use name lists and gazetteers to recognise references to known persons and places, and we use lexical patterns to identify references to new persons and organisations. The lexical patterns found (e.g. 'former Yugoslav president') are stored in the database as 'Titles' because they provide useful information about each person. Disambiguation heuristics are used to distinguish between places and persons with the same name (e.g. 'Victoria') and to disambiguate between various places with the same name (such as the 45 places called 'Paris' worldwide). For details, see [10] and [13]. For each place name, the co-ordinates are identified and used to generate maps. We have developed tools to identify date expressions in text [7], but this information is not yet integrated with the rest of the system.

## 2.2. Keywords and country score

For each document, we also extract a ranked list of keywords and their keyness (the degree with which the keyword is outstandingly frequent in the text), using the statistical log-likelihood test ([3]). The reference word frequency list used was built using the same text type, i.e. large numbers of news texts.

Internally, each text is represented by a vector consisting of a long list of keywords and their keyness. This vector is enhanced by the normalised *country scores* of the text. Country scores are calculated by adding one counter each time the country or one of its cities is referred to, and by then normalising this text occurrence frequency with the average occurrence frequency of the country in the reference corpus (see [11] for details). Country scores are expressed as country ISO codes and their keyness (the log-likelihood value). Adding the country score to the list of keywords helps to distinguish clusters with contents that are similar, but distinct, such as election campaigns in different countries. Its representation by the ISO code furthermore helps to find related clusters in different languages (see Sect. 3.3).

## 2.3. Specialist terms

The lists of people, organisations, places and keywords give users a good idea of what the document is about. However, analysts are normally specifically looking for information regarding their own subject domain. It is therefore useful to additionally display a list of those subject-specific terms that were found in the document or in a group of documents (See Table 1). If users provide term lists, the system automatically checks for each of the terms whether it is found in the text and, in the positive case, lists the terms. This term list tells the users about the specific relevance of this document for their field of interest.

**Table 1. List of specialist terms from the field of nuclear non-proliferation found in a cluster of six news articles about North Korea.**

| TERM | FREQUENCY |
| --- | --- |
| nuclear | 39 |
| weapons of mass destruction | 12 |
| missile | 9 |
| uranium | 7 |
| plutonium | 7 |
| disarmament | 6 |
| IAEA | 4 |
| proliferation | 3 |
| atomic | 2 |
| biological | 1 |
| reprocessing | 1 |
| enrichment | 1 |
| nuke | 1 |
| scud | 1 |

## 2.4. Morphological variants

The lookup of terms and names of people or places in plain text will only be successful if the looked up term is morphologically identical with the string found in the text. For inflected languages, we therefore provide exhaustive suffix lists that will be appended to the search term in all combinations so that any of the morphological variations will be found. By doing this, we can ensure, for example, that the inflected Estonian place names *Londonit* and *New Yorgile* are successfully recognised and mapped to *London* and *New York*. See [14] for details.

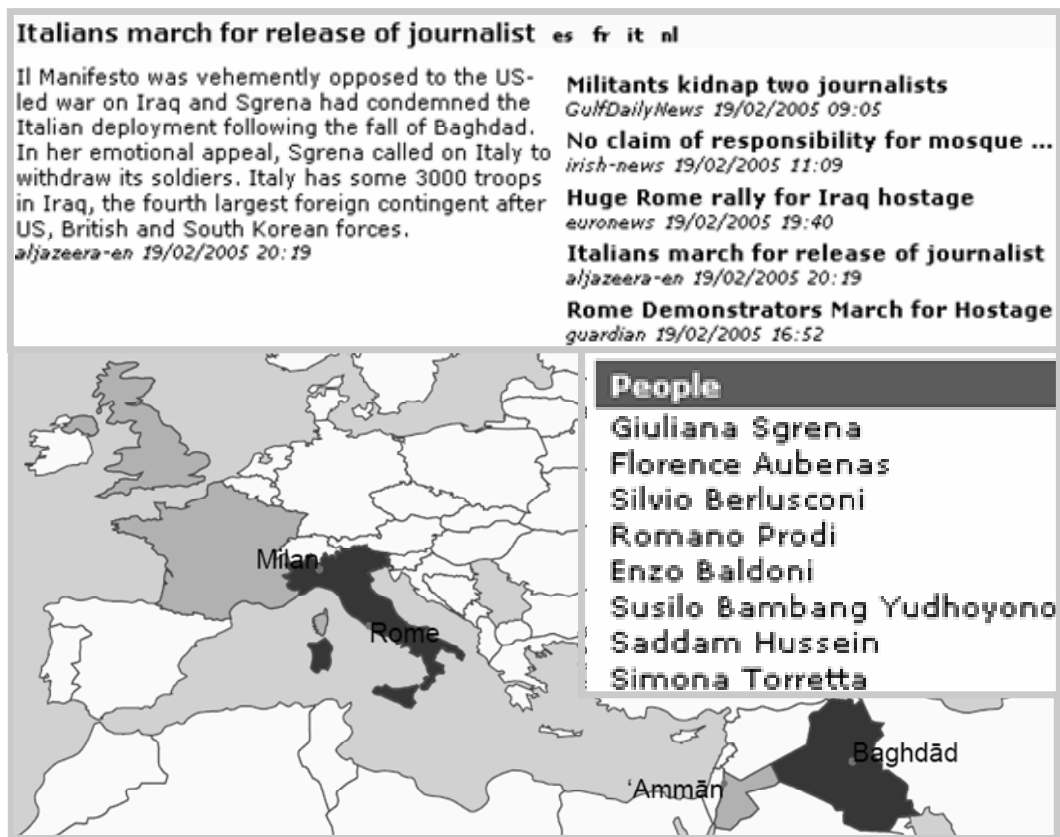

**Figure 1. English news cluster consisting of 5 documents, with automatically identified links to related news clusters in Spanish, French, Italian and Dutch. Hyperlinks on the titles allow users to read the individual articles; links on names allow them to find out more about the persons, about all news articles in which they are mentioned, etc.**

To make the lookup process of the hundreds of thousands of names, terms and their variants computationally viable, the regular expressions are compiled into a Finite State Automaton using FLEX ([9]).

## 2.5. Eurovoc thesaurus classification

We also map the contents of the whole news cluster (see Sect. 3.1) onto the multilingual thesaurus Eurovoc ([5]). Eurovoc is a list of over 6000 hierarchically organised concepts that is used by many European parliaments for the manual indexing and classification of documents in their libraries. The wide-coverage thesaurus descriptors (class names) exist in one-to-one translations in over twenty languages. The automatic categorisation tool was trained on manually classified documents. The approach was to build a profile-based category-ranking tool, which considers a number of parameters that have been empirically optimised to achieve maximum results. Details of the system are described in [12]. State-of-the-art categorisation methods were not used due to the extraordinary complexity of the unbalanced, multi-label classification data and the size of the training set. Depending on the language, up to 80000 documents had been manually classified according to over 5000 descriptors with an average of 5.6 descriptors per text.

The result of the thesaurus mapping is a vector of the 100 most relevant Eurovoc descriptors and their numerical identifiers (selected out of the pool of 6000 Eurovoc descriptors), and their relevance. This vector is used to link similar news clusters across languages (see Sect. 3.3).

## 3. Clustering texts and linking clusters

### 3.1. Clustering and topic detection

In order to group related news articles into a cluster, we build a hierarchical clustering tree (dendrogram), using an agglomerative algorithm ([8]). This tree is built by first calculating a pairwise similarity between all document pairs, using the cosine on the document vectors described in

Sect. 2.2. The most similar pair of vectors is then combined into a new vector and a new representation is built for this sub-cluster, by merging the keywords and by averaging their keyness. This new node will henceforth be treated like a single document, with the exception that it will have twice the weight of a single document. The hierarchical, binary clustering continues in this way until all documents are part of the cluster. The resulting dendrogram will have clusters of articles that are similar, and a list of keywords and their weight for each cluster. The degree of similarity for each cluster is shown by its intra-cluster similarity value, or cluster cohesiveness.

In a next step, we search the dendrogram for the major news clusters of the day (*topic detection*), by identifying all sub-clusters with an empirically derived minimum intra-cluster similarity of 50%. For further details, see [11].

### 3.2. Cluster Representation

For each cluster, we choose the article that is most similar to the cluster's centroid and we use its title as the title for the whole cluster. For each cluster, we furthermore display the major keywords and the sum of all names and terms found in the individual articles of the cluster (see Fig. 1). We also generate a zoomable and interactive SVG map ([4]) showing the geographical coverage of the articles in this cluster. By moving the mouse over a country or place name, the occurrences of the name are shown in their context (see Fig. 2).

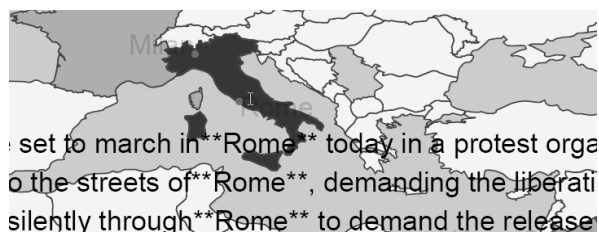

**Fig 2: Geographical map using SVG, allowing to zoom and to display the context of places.**

### 3.3. Linking clusters over time and across languages

As the whole cluster has a representation similar to that of each individual article, we can display the keywords for the cluster and we can also calculate the similarity between this cluster and clusters of other collections. For instance, in our daily news analysis, we compare each cluster with the clusters of the previous days so as to show the relationships between clusters over time (*topic tracking*).

We also link similar clusters across languages. When calculating the similarity between clusters in different languages, we use three vectors as ingredients, with a relative weighted impact of 70%, 20% and 10%, respectively, on the overall cross-lingual cluster similarity (see [11] for details): (1) We map the cluster onto the multilingual classification thesaurus Eurovoc (see Sect. 2.5); (2) We produce a vector representing the geographical coverage, using the weighting described in Sect. 2.2; (3) We calculate the similarity between the monolingual vector representations of the clusters (not considering the enhancement with the country score).

The reason for using these three ingredients is the following: The Eurovoc vector representation places each cluster into a generic conceptual space. The geographical vector ensures that similar stories such as elections are well distinguished from one country to the other. The monolingual vector, consisting of the keywords of the cluster, mainly contributes to the cross-lingual cluster similarity if some strings such as names or cognates are the same across languages. It is planned to replace the third vector by two more specific vectors: one consisting of person and organisation names and the other cluster consisting of dates and numbers.

### 3.4. Access to the NewsExplorer

The results of the daily clustering, cluster linking and information extraction have been integrated into the JRC's *NewsExplorer* system, which can be accessed at the address http://press.jrc.it/NewsExplorer (Fig. 3). The zoomable maps and some other details are not yet available on this public web site.

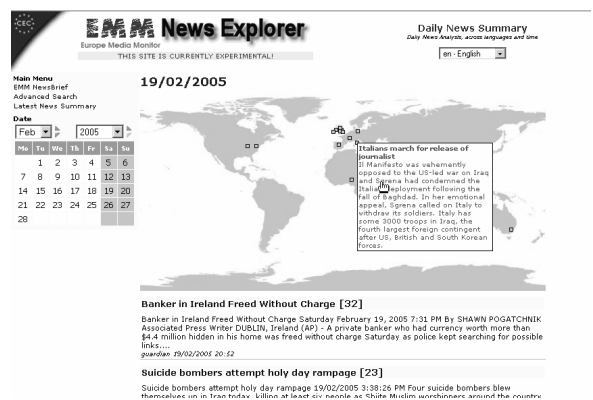

**Fig 3: Public NewsExplorer web site.**

## 4. Learning relations over time

### 4.1. Name variants

Due to differing transcription standards, foreign names are frequently spelled differently across languages, and even within the same language. An example is former Iraqi Interim Prime Minister *Iyad Allawi*, for whom we found variants like *Iyad Alaui*, *Ajad Allawi*, *Ijad Alawi*, *Illyad Allaoui*, and many more. We often find name variants for the same person within one cluster. We therefore apply an approximate name matching algorithm on all names found in the same cluster to identify those name strings that are likely to refer to the same person. The algorithm calculates the cosine of the letter bigram and trigram frequency lists of each name pair, automatically groups those whose similarity is above the empirically determined threshold of 0.7, and stores them under the same numerical name identifier. Name variants not automatically identified (e.g. *Joseph Ratzinger* and *Benedict XVI*) or wrong matches (like those between the similar names *George Bush jr.* and *George Bush sr.*) can be corrected interactively. For some Arabic and Russian names, our database has collected over sixty variants (see [13]).

### 4.1. Entity relations

The NewsExplorer system currently analyses news in English, German, French, Spanish, Italian, Dutch, Slovene and Estonian. It produces an average of between 20 and 130 clusters per language and per day. For each of the clusters, a number of people, organisations, country scores and keywords are identified and stored in a relational database. Over time, this database builds up statistics on the frequency of co-occurrence of any of these entities in the same cluster. As the system has been running for over a year, a simple database query can tell how often people have been mentioned in individual articles or in clusters. It can also give information on which names are frequently mentioned together with which others. As the names of some famous politicians are mentioned disproportionately often, a weighting algorithm suppresses personalities that are associated with many different other names, and it picks up name pairs that are mainly mentioned together and that are not associated with too many other names. For details on learning relations between people and on the weighting mechanism, see [13]. Table 2 shows the top of the list of weighted associated names for the Formula-1 driver *Michael Schumacher*.

**Table 2. List of names associated with *Michael Schumacher*, as of 28.04.2005.**

| NAME | COMM-CLUST | WGHT |
|---|---|---|
| Kimi Raikkonen | 142 | 1.10 |
| Takuma Sato | 126 | 1.0 |
| Giancarlo Fisichella | 147 | 1.0 |
| Rubens Barrichello | 246 | 0.99 |
| Fernando Alonso | 234 | 0.95 |
| Felipe Massa | 76 | 0.94 |
| Jarno Trulli | 179 | 0.91 |
| Jenson Button | 189 | 0.90 |
| Nick Heidfeld | 143 | 0.89 |
| Juan Pablo Montoya | 86 | 0.88 |
| David Coulthard | 137 | 0.88 |
| Jean Todt | 58 | 0.83 |
| … | | |

The comparison of Column 2 (number of clusters in which both names occur) versus Column 3 (weighted association measure) shows that the number of clusters in which both names co-occur is not the most significant feature, because a number of further features are considered in the weighting.

## 5. Exploring the document collection

The information collected in the processes described in the previous sections is stored in a relational database. Information that needs to be accessed quickly (e.g. daily information in the *NewsExplorer* application) is additionally exported into XML files. The user interfaces can be viewed with common web browsers.

For a document collection that can consist of all the news of the day or of any other selection, users can explore the available information via several entry points. The first and major entry point is the list of clusters, which gives the users an overview of what can be found in the collection. For each cluster, an individual HTML page can be created on the fly, containing a list of all individual articles belonging to the cluster, a geographical map, information about the cluster's keywords, person and organisation names, and lists of clusters that are related to the chosen one (see Fig. 1). These linked clusters are either those that were published in the previous days about the same story, or they are the ones that talk about the same story in other languages.

Each of the individual articles has a hyperlink to the URL where it was originally found so that the user can read the full article (if it is still available). When clicking on any of the keywords or on any of the place names, the system displays a list of all clusters for which this keyword or place was identified. As the keywords are language-dependent, the list of related articles will be in the same language as the original document, but cluster lists for place names can be in any of the languages, because the geographical information (place ID, latitude and longitude) is language-independent.

When clicking on any of the person or organisation names, additional information about this name will be displayed, where available, as shown in Fig. 4.

The item 'titles' contains the list of lexical patterns in the various languages that helped us identify the name. If users do not know a person and no Wikipedia encyclopaedia entry is available, the list of titles, associated keywords and countries gives them a rather good idea of the context of the person.

Instead of starting navigating the document collection via the list of clusters, users can also use a search form to enter a name, keyword or country. If the entered name is not found in the database of names and their variants, the approximate name matching process will be started to find names spelled similarly to the one sought. For any keyword or place name entered, the user will be presented with the same links as shown in Table 3, except obviously the Wikipedia entries and the name-related links ('titles', 'name variants' and 'weighted associated names'). The navigation and exploration tool exploits the full power of the relational database whose knowledge about relations has been built up over time.

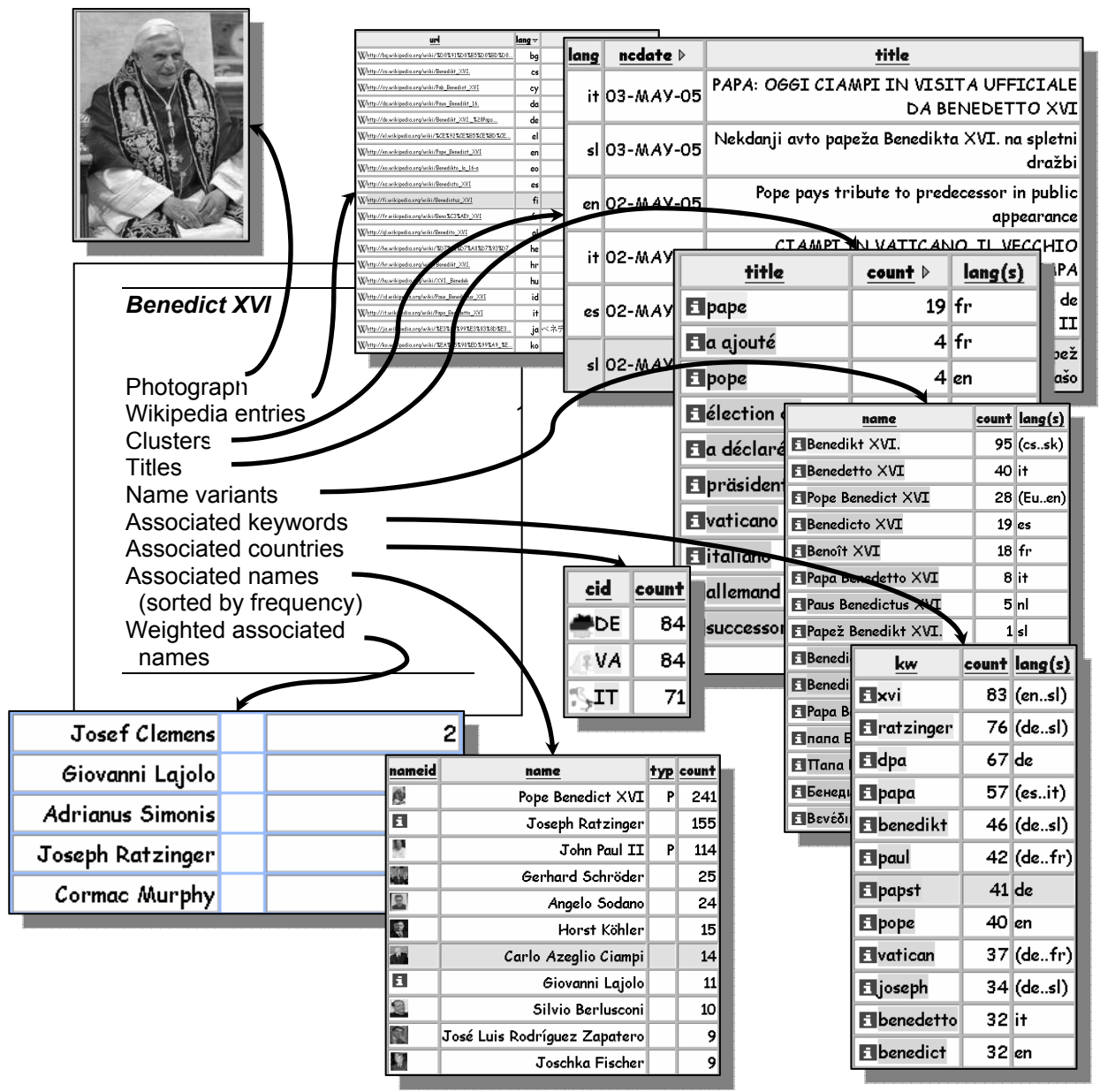

**Fig. 4. Information available for persons in the NewsExplorer database.**

## 6. Conclusion and future work

We have developed a multi-component system for multilingual news analysis that can be used by analysts who frequently have to look through large news collections to find information about their specific field of interest. The proposed system first organises the whole collection into groups of similar articles by applying state-of-the-art clustering methods. It then analyses the collection, cluster by cluster, by extracting and displaying information such as named entities and occurrences of controlled vocabulary specialist terms. The system stores all the extracted information in a relational database. As the system analyses vast amounts of news articles every day, the database contains a lot of relational information that the user can access. The acquired database knowledge includes information on which persons or organisations are frequently or mainly mentioned in the same news stories, which keywords and countries are associated with each person, what spelling variants the press uses for the same person, etc. Users can browse the document collection and the information stored in the database.

The system is in an advanced prototype phase, but some of the current functionality is already available on the public web site of the JRC's *NewsExplorer*.

It is planned to extend the functionality and the language coverage of the NewsExplorer. Eventually, news from all 20 official EU languages should be included. The system will be improved by extracting further entities such as dates and currency expressions, and by using these plus extracted person names to improve the linking of related news clusters in different languages.

We also plan to apply the tool to the analysis of document collections of different types, such as in-house collections of documents, the results of search engine queries or the results of subject-specific or site-specific web-crawling. This application to new text types brings new challenges, such as the need for document format conversion and the need to tune the named entity recognition software to other text types.

Finally, we would also be interested in detecting the more specific relationships between the entities found in the news – which leads us to event scenario template filling – and we would like to gather information about the sentiment expressed in the news report ('How' the event is reported).

## 7. Acknowledgements

We thank the JRC's Web Technology sector, and especially Clive Best, Erik van der Goot, Kenneth Blackler and Teofilo Garcia, for the access to the *Europe Media Monitor* news data and for their support. We furthermore thank the many persons who have contributed over time to develop the existing text analysis tool set and to adapt it to so many languages.